%% file: main.tex
\documentclass[conference,letterpaper]{IEEEtran}
\IEEEoverridecommandlockouts
\usepackage{cite}
\usepackage{amsmath,amssymb}
\usepackage{graphicx}
\usepackage{booktabs}
\usepackage{url}
\usepackage{xspace}
\usepackage[nospread,noshrink]{flushend}
\graphicspath{{figures/}}
\input{data/curation_values.tex}
\input{data/split_half_values.tex}
\newcommand{\fhalf}{\ensuremath{F_{0.5}}\xspace}
\newcommand{\nfmr}{\ensuremath{\mathrm{N\mbox{-}FMR}}\xspace}
\newcommand{\tsel}{\ensuremath{\tau_{\mathrm{sel}}}\xspace}
\newcommand{\tabs}{\ensuremath{\tau_{\mathrm{abs}}}\xspace}

\begin{document}
\title{Curating Merchant-Matching Training Data\\with Two Confidence-Gated Local LLM Judges}

\author{
\IEEEauthorblockN{Donghao Huang\textsuperscript{1}, Jinling Pei\textsuperscript{2},
 Zhaoxia Wang\textsuperscript{2,*}}
\thanks{*Corresponding author: zxwang@smu.edu.sg}
\IEEEauthorblockA{\textsuperscript{1}Research and Development, Mastercard, Arlington, VA, USA\\
\textsuperscript{2}School of Computing and Information Systems, Singapore Management University, Singapore\\
}
\IEEEauthorblockA{
donghao.huang@mastercard.com, jinling.pei.2026@engd.smu.edu.sg, zxwang@smu.edu.sg
}
}
% \author{\IEEEauthorblockN{Anonymous Submission}}
\maketitle

\begin{abstract}
Merchant matching resolves a noisy payment descriptor to a retrieved merchant entity or returns no match. A key challenge in curating reliable training labels is distinguishing teacher abstention from genuine evidence that no acceptable entity exists: false no-match labels can contaminate pseudo-labeled training data, while overly conservative labeling reduces coverage.
%2. OBJECTIVE / METHOD / SOLUTION TO THE PROBLEM
This work investigates whether agreement between two local large language model judges can improve pseudo-label reliability for merchant matching. The approach accepts a label only when the two judges agree and uses ordered selection and abstention thresholds to guarantee disjoint positive and negative label sets.
%3. RESULTS
Retrospective replay on \cvBenchN{} expert-annotated queries shows that higher selection thresholds can improve positive-label purity, whereas higher abstention thresholds increase false no-match labels. At thresholds $(\cvOpPosTau,\cvOpNegTau)$, Muse Glimmer 30B and Gemma 4 31B jointly label \cvDualTotalN{} queries (\cvDualCoverage\% coverage) at \cvDualTotalPurity\% purity, with positive and negative purities of \cvDualPosPurity\% and \cvDualNegPurity\%, respectively.
Overall purity exceeds that of either constituent model at the same thresholds by more than two percentage points, albeit with lower coverage.
% [R1-W1][R1-W4] New sentence: the held-out check of the operating point.
A repeated split-half check shows that this inherited operating point coincides, within one grid step, with the optimum of a stated objective, and that threshold-selection optimism is small (\cvShOptimism{} percentage points of total purity on held-out halves).
% [CR] "--" replaces a literal en dash so the source is ASCII-safe.
A symmetric threshold of \cvOpPosTau{} adds \cvSymHighExtraWrongNeg{} erroneous no-match labels, while \cvDualWrongNeg{} false abstentions persist even when the confidence threshold is removed. Across \cvEffContrastN{} matched within-model comparisons, higher reasoning effort yields no clear \fhalf{} gain but increases median latency by \cvEffLatencyMinX--\cvEffLatencyMaxX{} times.
%
%4. CONCLUSIONS
These results reveal asymmetric error behavior between positive and negative pseudo-labels, motivating separate thresholding and auditing rather than a single symmetric confidence criterion. They also indicate that increased reasoning effort does not necessarily improve labeling quality under the evaluated conditions.
%
%5. INSIGHT / FUTURE WORK/ SIGNIFICANCE FOR OTHER RESEARCHERS (If the abstract is too long, this part can be omitted.)
% [R1-W2] Part 5 filled in: the abstract now states what the study does not
% establish, so the downstream-utility limitation is visible up front.
The study establishes label purity, not student utility: fresh-data curation and student fine-tuning remain necessary to show downstream value.

\end{abstract}


\section{Introduction}
Payment systems receive noisy records such as \texttt{EXMPL MKT 17} with
\texttt{123 EXAMPLE ST} (synthetic). A judge selects a retrieved merchant or
abstains. Wrong selections contaminate merchant-level aggregation; abstentions
% [CA-1] Name the metric here and point to the justification in Section III.
leave records unenriched, motivating the precision-weighted \fhalf{} objective
explained in Section~\ref{sec:task}.

The system studied here uses GPT OSS 120B~\cite{openai2025gptoss} on a managed
endpoint. Our downstream goal is to curate fresh training data and fine-tune
smaller local judges, including GPT OSS 20B and Gemma 4 12B, using low-rank
adaptation (LoRA)~\cite{hu2021lora}. Expert annotation is costly, and sensitive
merchant records may include sole proprietors. Local teachers can generate
labels within the operator's environment, but which outputs are reliable enough
to retain?

Agreement has two meanings: selecting the same merchant proposes a positive
label; joint abstention may reflect an absent candidate, insufficient evidence,
or a shared mistake. A confidence gate also converts tentative selections into
abstentions. Raising one shared threshold can therefore clean positive labels
while making negatives noisier.

This paper investigates such a distinction with two local judges, Muse Glimmer 30B and Gemma
4 31B, on a frozen expert-annotated benchmark. In this empirical study, we:
\begin{itemize}
\item formalize separately gated agreement for positive and negative labels,
including a threshold-ordering condition that guarantees no contradictory labels;
\item measure purity and coverage over a threshold grid, with a controlled
symmetric-gate ablation and comparisons to each constituent judge;
% [R1-W1][R1-W4] New contribution bullet for the held-out check.
\item state an explicit selection objective and check the inherited operating
point against it with repeated split-half selection and held-out scoring;
\item audit shared false abstentions and the deferred rows to identify what
threshold changes can and cannot repair in the stored outputs;
\item quantify within-model reasoning-effort trade-offs using paired quality
comparisons and measured latency to inform teacher configuration.
\end{itemize}
We replay existing outputs without student training or new production labeling.
% [R1-W1] Sharpened: the split-half check bounds only the threshold part of the
% selection optimism; model and prompt selection remain in-sample.
Model and threshold selection reuse this benchmark, so estimates are
retrospective. The split-half check in Section~\ref{sec:splithalf} quantifies
the threshold-selection part of that optimism; the model and prompt choices
remain in-sample.

\section{Related Work}
\paragraph*{Entity matching}
Record linkage combines noisy evidence under asymmetric error
costs~\cite{fellegi1969theory,christen2012data}. LLM-based approaches include
pairwise classification, example selection, and
fine-tuning~\cite{peeters2025entity,fan2024costeffective,steiner2025finetuning}.
Wang et al.~\cite{wang2025match} compare matching, comparing, and selecting
strategies and combine them in ComEM. CaRL-EM learns cost-aware control of
matching operators~\cite{guo2026carl}. Our unit of curation is a query and its
candidate pool, with a selected record or a no-match label.

\paragraph*{Task-dependent reasoning}
Huang and Wang~\cite{huang2026task} evaluate 504 configurations across binary
sentiment, five-class sentiment, and 27-class emotion recognition, studying both
reasoning-focused distillation and thinking-mode activation. Their aggregate
results favor base/non-thinking models on the first two tasks, while reasoning
variants improve emotion recognition at higher latency. Huang et
al.~\cite{huang2027scale} compare 46 configurations on eight public clean-clean
entity-matching benchmarks under a fixed ComEM protocol. They find favorable
quality--cost trade-offs at low effort for GPT-5, but higher $F_1$ at high effort
for GPT OSS 120B and 20B, with heterogeneous thinking effects across Claude
models. These studies motivate evaluating reasoning jointly with task, model,
and computational cost, rather than assuming that more reasoning improves
matching quality.

\paragraph*{LLM annotation and distillation}
LLMs have been evaluated as text annotators~\cite{gilardi2023chatgpt} and as judges
of generated responses~\cite{zheng2023judging}. More directly, Steiner and
Bizer~\cite{steiner2026labeling} study LLM-labeled entity-matching pairs, varying
pair selection, teachers, label post-processing, and student models across five
public benchmarks. Their work demonstrates downstream distillation; ours examines
label quality in a proprietary, listwise matching task, particularly the risk of
interpreting agreement on abstention as evidence for a negative label.

\paragraph*{Weak supervision and selective prediction}
Snorkel combines noisy labeling functions while accounting for their dependencies
and accuracies~\cite{ratner2017snorkel}. This paper uses a simpler deterministic agreement
% [CR] grammar: "assess" -> "assesses" (subject is "this paper").
rule and assesses it directly against expert labels. Selective prediction studies
the trade-off between coverage and risk~\cite{geifman2017selective}; here the risk
is label error among retained examples. A crucial distinction is that a labeling
function's abstention ordinarily withholds supervision, whereas our policy converts joint judge abstention into a proposed no-match label. That
conversion must be audited. The intended downstream use is teacher-generated
hard-label supervision, related to knowledge distillation~\cite{hinton2015distilling},
but student utility remains unmeasured.

% [R1-W6] New paragraph: the reviewer asked for recent literature on evidence
% acquisition for multimodal financial reasoning. Deng et al. (Information
% Fusion, 2026) is the directly relevant work; the paragraph relates it
% honestly to this study (fixed pool, no evidence gathering) and points to the
% deferred/audited rows as the natural place for such acquisition.
\paragraph*{Evidence acquisition in financial reasoning}
In multimodal financial reasoning, Deng et al.~\cite{deng2026sea} attribute many
failures of multimodal LLMs to incorrect intermediate evidence rather than to
the final inference step, and propose step-wise evidence acquisition (SEA), in
which cooperating agents decompose a task, retrieve targeted evidence, and reason
over it iteratively. The judges studied here reason once over a fixed retrieved
candidate pool and acquire no further evidence; agreement between them is a
retention signal, not an evidence-gathering step. The shared false abstentions
in Section~\ref{sec:residual} are precisely cases where the fixed pool's
evidence was insufficient or misread by both judges, which makes targeted
evidence acquisition for deferred and audited rows a complementary direction.

\section{Task and Experimental Setup}
\label{sec:task}
\paragraph*{Task and gate}
For query $i$, let $C_i$ be a retrieved pool of 19--21 candidate records and
$Y_i\subseteq C_i$ the expert-accepted records. $Y_i=\varnothing$ means no
acceptable record is present, not that the merchant does not exist elsewhere.
The judge returns a \texttt{row\_id}, a name-match flag $m$, name confidence $m_c$,
and field-agreement flags. The evaluator accepts a non-null selection only if
$m=1$, $m_c\geq\tau$, and the five location flags (street, ZIP, city, state,
country) are each 1 or null. A valid abstention can contain only
\texttt{"row\_id": null}. Thus rejection by the gate and explicit abstention
produce the same deployed action. Both prompt versions state a $.60$ confidence
cutoff; the deployed evaluator uses $\tau=\cvDeployedTau$. Our sweep includes both.

\paragraph*{Benchmark and demonstrations}
The frozen evaluation set contains \cvBenchN{} expert-annotated queries:
\cvBenchPos{} with at least one acceptable candidate and \cvBenchNeg{} with none.
An additional 40 queries form a disjoint demonstration pool. Gemma uses a fixed
20-example prefix drawn from that pool, with expert target labels and
agreement-filtered structured teacher verdicts. Query IDs do not overlap between
the pool and evaluation set, but entity disjointness is not guaranteed.
Accepting any member of a multi-record gold set counts as correct.
% [R1-W5] The reviewer asks for the annotation procedure and inter-annotator
% agreement. Nothing verifiable could be added from the artifact, so the
% disclosure is kept and its consequence for the purity metrics is made
% explicit. 
The original annotation procedure and inter-annotator agreement are not
documented in the available artifact, limiting independent assessment of label
reliability. All purity figures below are therefore agreement with these expert
labels, including their multi-record gold sets, rather than with an
independently adjudicated truth.

\paragraph*{Configurations and serving}
This work pairs Muse Glimmer 30B~\cite{meta2026museglimmer} at zero-shot, low effort, prompt
v1 with Gemma 4 31B~\cite{gemma2026report} at $k=20$, prompt v2. v1 uses
street-conditional rule templates; v2 unifies and revises the rules, including
ambiguity handling and tie-breaking. These configurations were selected using
previous comparisons on this same benchmark. Muse has higher selection precision
and lower \nfmr, while Gemma has higher match recall
(Table~\ref{tab:judges}). This motivates the pairing but does not prove error
independence or that this pair is optimal among other pairs.
% [R1-W3] Made the prompt/demonstration confound explicit at the point where
% the configurations are introduced, not only in the Limitations section.
Because the two judges differ in prompt version and in demonstrations (zero-shot
versus $k=20$), differences between them, and the pair's gains over either one,
cannot be attributed to model family alone; the pair is evaluated as a
deployed configuration, not as a controlled model comparison.

The local runs use an Apple M5 Max workstation with 128\,GB unified memory;
Muse and Gemma use bf16 weights served through Ollama. The recorded local software
inventory spans Ollama 0.32.0--0.33.1 across batches, but versions and weight
digests were captured after inference rather than per run. GPT OSS 120B
(zero-shot, low effort, v1, local MXFP4) and Claude Sonnet 4.5
(zero-shot, low effort, v1, API)~\cite{anthropic2025sonnet} are references.
The deployed medium-effort run is a separate quality reference. Latency is serial
model-call time, not a measurement of concurrent production serving or annotation
cost. Local curation requires both constituent calls per query.
Low denotes enabled, provider-specific reasoning; off denotes disabled thinking.
Section~\ref{sec:effort} examines the quality and latency implications of these
settings where matched effort runs are available.

\input{data/table_judges.tex}

\paragraph*{Judge metrics}
Correct matches (CM), wrong candidates on matchable rows (WC), spurious matches
on no-match rows (SM), and missed matches (MM) define
\begin{align}
P &= \frac{\mathrm{CM}}{\mathrm{CM}+\mathrm{WC}+\mathrm{SM}}, &
R_m &= \frac{\mathrm{CM}}{\mathrm{CM}+\mathrm{WC}+\mathrm{MM}},\\
F_{0.5} &= \frac{1.25PR_m}{0.25P+R_m}, &
\mathrm{N\mbox{-}FMR} &= \frac{\mathrm{SM}}{|\{i:Y_i=\varnothing\}|}.
\end{align}
A wrong-candidate selection counts against both precision and recall.
% [CA-1] Brief justification of F0.5 over F1 (co-author request). Three
% reasons, in one short passage: asymmetric error costs in this application,
% the beta=0.5 weighting (precision counts twice as much as recall), and
% consistency with the deployed judge's own acceptance objective. N-FMR is
% named as the companion metric that exposes what a precision-weighted score
% can hide. Mirrors the rationale used in paper/eacl.
We report \fhalf{} rather than $F_1$ because the two error types carry
asymmetric costs here: a wrong or spurious selection attaches a record to the
wrong merchant and propagates into merchant-level aggregates, whereas a missed
match leaves the record unenriched and can be retried or reviewed later.
$F_\beta$ with $\beta=0.5$ weights precision twice as much as
recall~\cite{vanrijsbergen1979information}, and it is also the acceptance
objective of the deployed judge, so every configuration is compared under the
production criterion rather than a symmetric one. Because a precision-weighted
score can still conceal aggressive selection on no-match rows, \nfmr{} is
reported alongside it. These metrics characterize the teachers; the curation
outcomes below are label purity and coverage.

\section{Separately Gated Agreement}
\label{sec:pipeline}
\paragraph*{Policy}
Let $g_\tau(j,i)$ denote judge $j$'s selected record after the gate, or $\bot$
for explicit abstention or gate rejection. For fixed stored outputs, raising
$\tau$ can only change a selection to $\bot$; it cannot create or change a
selection. With judges $A$ and $B$, define
\begin{align}
\mathcal{P}(\tsel) &= \{i:g_{\tsel}(A,i)=g_{\tsel}(B,i)\neq\bot\},\\
\mathcal{N}(\tabs) &= \{i:g_{\tabs}(A,i)=g_{\tabs}(B,i)=\bot\}.
\end{align}
Each row in $\mathcal{P}$ receives the agreed record ID. Each row in
$\mathcal{N}$ receives a no-match pseudo-label. The remaining rows form a review
set $\mathcal{R}$. Exact record-ID agreement is conservative: two different
selections are deferred even when both happen to belong to $Y_i$.
The policy does not consult $Y_i$ when producing labels.

\paragraph*{Disjointness guarantee}
The condition $\tabs\leq\tsel$ guarantees
$\mathcal{P}\cap\mathcal{N}=\varnothing$ for all outputs: both judges selecting
at \tsel{} must also select at the lower \tabs{} by monotonicity.
If $\tabs>\tsel$, two judges selecting the same record with confidence in
$[\tsel,\tabs)$ place that row in both sets. Inversion need not cause overlap
on every dataset, but loses the general guarantee; here $(.80,.86)$ produces
\cvInvertedOverlap{} contradictory labels. We enforce the ordering; an explicit
conflict-resolution rule could instead define a different policy.

\paragraph*{What the thresholds control}
Raising \tsel{} makes $\mathcal{P}$ shrink; raising \tabs{} makes
$\mathcal{N}$ grow. These set relationships are structural. The corresponding
\emph{purities are not necessarily monotone}, because the correctness of rows
added or removed depends on the data. In particular, \tabs{} is a threshold on
\emph{selection} confidence, not a confidence score for the no-match label.
A gate-induced abstention cannot by itself certify absence of a valid candidate.

\paragraph*{Purity, coverage, and uncertainty}
Positive purity is the fraction of retained selections in $Y_i$; negative purity
is the fraction of retained no-match labels with $Y_i=\varnothing$. With correct
label counts $c_P$ and $c_N$, total purity and coverage are
\begin{equation}
\pi=\frac{c_P+c_N}{|\mathcal{P}|+|\mathcal{N}|},\qquad
\kappa=\frac{|\mathcal{P}|+|\mathcal{N}|}{\cvBenchN}.
\end{equation}
Total purity is sensitive to the mix of positive and negative labels, so this paper
reports both components. Empty sets have undefined purity. Wilson 95\% intervals
summarize individual proportions. Purity differences between policies use
\cvBootstrapN{} paired bootstrap resamples of all evaluation rows (seed 42),
recomputing each policy's numerator and selected-set denominator. These are
pointwise, retrospective intervals; they do not account for configuration search,
entity clustering, or repeated inference variability.

\paragraph*{Replay and failure handling}
In this research, we sweep each threshold from \cvTauGridLo{} to \cvTauGridHi{} in $.01$ steps
without re-inference. All scored runs contain \cvBenchN{} parseable,
error-free outputs. The analysis checks row alignment, reproduces recorded
decisions at the deployed gate, and verifies monotone set membership and
disjointness over the valid grid. A failed request or unparseable response must
be retried or reviewed, never treated as a negative training label; the analyzer
rejects such inputs. Input hashes and aggregate tables document the replay.

\section{Results}
\label{sec:results}
\subsection{Reasoning Effort and the Quality--Latency Trade-off}
\label{sec:effort}
\begin{figure*}[t]
\centering
\includegraphics[width=.996\textwidth]{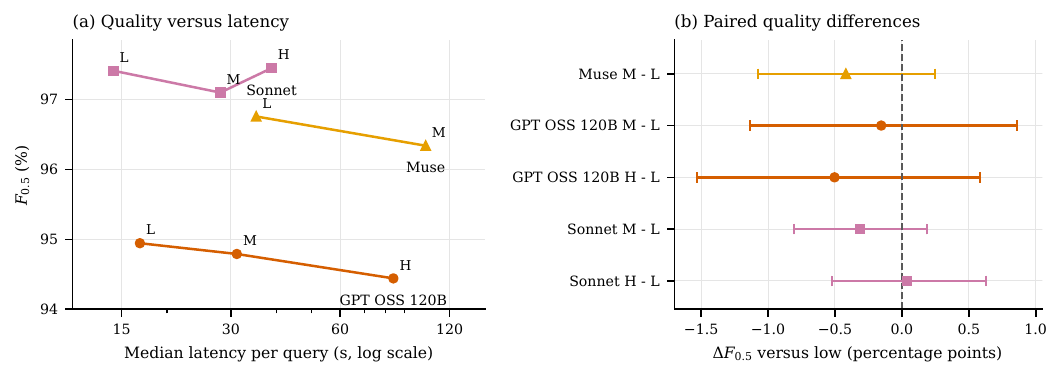}
\caption{Reasoning effort on the frozen benchmark. (a) Each trajectory connects
low (L), medium (M), and, where available, high (H) effort at $\tau=.70$.
Muse and GPT OSS 120B run locally; Sonnet uses an API. Latency comparisons are
within each serving backend. (b) Paired changes in \fhalf{} relative to low
effort; horizontal bars are pointwise 95\% bootstrap intervals. All five include
zero. Only complete runs are plotted; Gemma has no matched thinking-on run.}
\label{fig:effort}
\end{figure*}

This work compares \cvBenchN{}-query Muse, local GPT OSS 120B, and Sonnet runs, fixing
the model, backend, zero-shot v1 prompt, and $\tau=.70$ within each family.
Provider-specific effort settings do not imply equal compute across models.
Figure~\ref{fig:effort} relates \fhalf{} to median latency and shows
medium/high-minus-low quality differences with paired 95\% percentile intervals
(\cvEffBootstrapN{} row resamples, seed 42).

Relative to low effort, \fhalf{} changes by
$\cvEffMuseMediumDelta$ percentage points for Muse medium,
$\cvEffOssMediumDelta/\cvEffOssHighDelta$ for GPT OSS 120B medium/high,
and $\cvEffSonnetMediumDelta/\cvEffSonnetHighDelta$ for Sonnet medium/high.
Median latency increases by $\cvEffLatencyMinX$--$\cvEffLatencyMaxX\times$.
Muse medium improves precision but loses recall, illustrating why additional
reasoning need not improve the precision-weighted aggregate.
Even Sonnet's small high-effort gain has interval $\cvEffSonnetHighCi$ at
$\cvEffSonnetHighLatencyX\times$ latency.

All \cvEffContrastN{} intervals include zero: higher effort offers no clear
\fhalf{} gain here, supporting low effort when latency matters. This establishes
neither equivalence nor optimal curation purity. Muse high has only a latency
screen; Gemma lacks a matched on/off comparison. These results do not establish
an advantage of disabling reasoning. The same retrospective uncertainty
limitations apply as for curation.

The GPT OSS 120B point estimate differs in direction from the $+1.99$-point
low-to-high $F_1$ gain reported by Huang et al.~\cite{huang2027scale}.
Their study averages $F_1$ across eight clean-clean datasets after ranking
candidates down to four; ours evaluates confidence-gated \fhalf{} on noisy
merchant queries with 19--21 candidates. Their Sonnet comparison toggles
thinking on/off, whereas ours varies effort with thinking already enabled.
These differences limit direct transfer of an effort recommendation between
the protocols. The task dependence observed by Huang and
Wang~\cite{huang2026task} further motivates such evaluation, but our single
benchmark does not isolate task complexity or establish over-deliberation as
the cause of the observed quality changes.

\subsection{Curation Purity and Coverage}
\begin{figure*}[t]
\centering
\includegraphics[width=\textwidth]{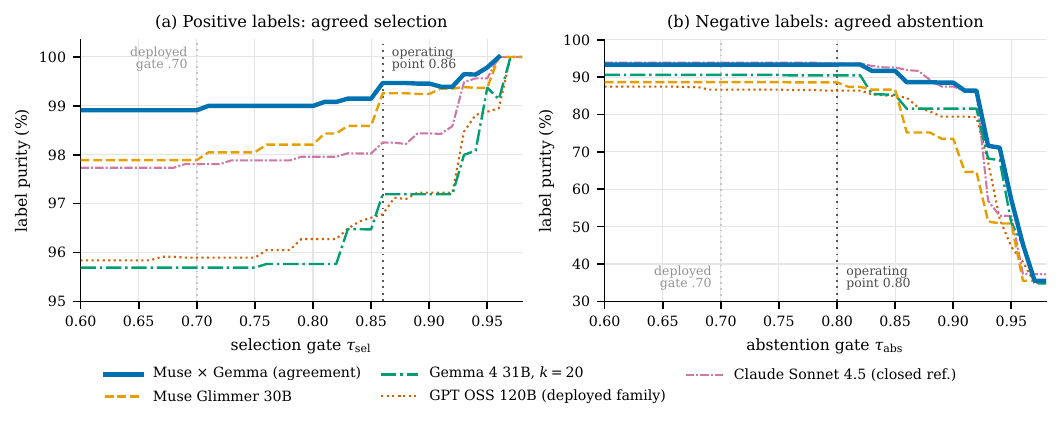}
\caption{Purity versus the selection threshold (a) and abstention threshold (b).
The panels use different vertical scales. Positive-label purity generally rises
as coverage falls; negative-label purity deteriorates sharply at high thresholds.
The vertical rules mark the deployed gate and illustrative operating point.
Curves summarize observed outputs, not calibrated probabilities or a guarantee
of monotone purity. Undefined purity for an empty positive set is omitted.}
\label{fig:sweep}
\end{figure*}

\paragraph*{Purity responds differently by label type}
Figure~\ref{fig:sweep} shows the two distinct responses. At \tsel$=.70$, the pair
retains \cvDualPosNAtSeventy{} positives at \cvDualPosPurityAtSeventy\% purity.
At \tsel$=\cvOpPosTau$, positive purity is \cvDualPosPurity\%; at $.95$ it reaches
\cvDualPosPurityAtNinetyFive\%, but only \cvDualPosNAtNinetyFive{} positives
remain. Small reversals between these points confirm that monotone membership
does not imply monotone purity.

For negatives, purity is \cvDualNegPurityAtEighty\% at
\tabs$=\cvOpNegTau$ and \cvDualNegPurityAtNinety\% at $.90$. The corresponding
false no-match counts are \cvDualWrongNeg{} and \cvDualWrongNegAtNinety;
at the sweep's upper endpoint the count reaches \cvDualWrongNegMax.
Raising \tabs{} admits additional gate rejections into $\mathcal{N}$, including
matchable rows on which the judges selected at lower thresholds. The observed
negative-purity decline is explained by the labels of these newly admitted rows.

\input{data/table_operating.tex}

\paragraph*{Illustrative operating point}
% [R1-W4] Cross-reference added: the inherited point is now checked against a
% stated objective in the new held-out paragraph below.
This work analyzes $(\tsel,\tabs)=(\cvOpPosTau,\cvOpNegTau)$, inherited from earlier
exploration of these outputs. It is an illustrative trade-off rather than a
derived optimum; Section~\ref{sec:splithalf} states an explicit objective and
checks the point against it on held-out halves. Table~\ref{tab:operating} reports \cvDualPosN{} positives at
\cvDualPosPurity\% purity (Wilson 95\% interval \cvDualPosCi) and \cvDualNegN{}
negatives at \cvDualNegPurity\% (\cvDualNegCi). Together these are
\cvDualTotalN{} labels at \cvDualTotalPurity\% purity (\cvDualTotalCi), with
\cvDualCoverage\% coverage and \cvDualReviewN{} deferred rows. There are
\cvDualPosErrors{} incorrect positive labels and \cvDualWrongNeg{} incorrect
negative labels. The negative subset therefore contains most of the retained
label noise despite the high aggregate purity.

\paragraph*{Agreement versus individual teachers}
At the same gates, Muse alone produces \cvMuseTotalNAtOp{} labels at
\cvMuseTotalPurityAtOp\% purity, and Gemma produces \cvGemmaTotalNAtOp{} at
\cvGemmaTotalPurityAtOp\%. The pair's purity differences are
$+\cvMusePurityDelta$ percentage points versus Muse (paired 95\% interval
\cvMusePurityDeltaCi) and $+\cvGemmaPurityDelta$ versus Gemma
(\cvGemmaPurityDeltaCi). These gains describe policies selecting different
subsets; they do not isolate agreement at matched coverage. Both constituent
policies retain more examples, and both incur more false no-match labels
(\cvMuseWrongNegAtOp{} and \cvGemmaWrongNegAtOp{}, respectively).

To give the single teachers the benefit of threshold tuning, we also maximize
each one's total purity over the full valid grid, subject to retaining at least
\cvDualTotalN{} rows. Muse reaches \cvMuseBestPurity\% on \cvMuseBestN{} rows and
Gemma \cvGemmaBestPurity\% on \cvGemmaBestN{}. Thus neither reaches the pair's
purity while meeting its coverage on this grid. These are descriptive search
maxima, not exact coverage-matched comparisons or out-of-sample guarantees.

Sonnet alone yields \cvSonnetTotalNAtOp{} labels at
\cvSonnetTotalPurityAtOp\% purity. The pair-minus-Sonnet difference is
$+\cvSonnetPurityDelta$ points (\cvSonnetPurityDeltaCi). This interval does not
establish equivalence or superiority. The local pair offers a different
purity--coverage trade-off and avoids remote inference for new curation queries;
the Sonnet reference itself was obtained through an API.

\input{data/table_ablation.tex}

\paragraph*{Why separate the gates?}
Table~\ref{tab:ablation} holds the judge outputs fixed and changes only the
retention rule. Relative to symmetric $.70$ gating, the illustrative policy
removes \cvSymLowPosLoss{} positive labels, including
\cvSymLowPosErrorsRemoved{} errors. This is a substantial volume cost for a small
positive-purity improvement; symmetric $.70$ or $.80$ remains a reasonable
alternative when volume matters more.
Conversely, setting \tabs$=\tsel=\cvOpPosTau$ leaves the positive set unchanged
but adds \cvSymHighExtraNeg{} negative labels, of which
\cvSymHighExtraWrongNeg{} are wrong, reducing total purity to
\cvSymHighPurity\%. Separating the gates avoids those added negative errors.
A positive-only policy attains \cvDualPosPurity\% purity while supplying no
negative training examples. Choosing among these policies requires explicit
requirements for class balance, label noise, and review capacity.

\input{data/table_grid.tex}

\paragraph*{Joint behavior}
Table~\ref{tab:grid} shows a relatively flat total-purity region for
\tsel{} between $.80$ and $.90$ and \tabs{} at or below $.80$. This stability does
not imply constant coverage. Increasing \tsel{} to $.95$ further purifies the
positive subset yet reduces \emph{total} purity because the cleaner positives
become a smaller share of the retained set. The two component sets can be analyzed
separately, but choosing their joint operating point remains coupled by threshold
ordering and label composition. No single best threshold follows without a
specified objective.

% [R1-W1][R1-W4] NEW paragraph and table. The reviewer's decisive limitation was
% that the operating point is inherited and never validated out of sample. We
% (i) state an explicit objective, (ii) show that its full-benchmark optimum
% coincides with the inherited point within one grid step, and (iii) quantify
% threshold-selection optimism by repeated split-half selection and held-out
% scoring. The paragraph is careful to say what this does NOT do: it cannot
% remove the model/prompt selection that also reused this benchmark. All
% numbers are macros written by split_half_analysis.py (stored outputs only).
\input{data/table_split_half.tex}
\paragraph*{Held-out check of the operating point}
\label{sec:splithalf}
To test whether the inherited point survives a stated objective, we define one:
maximize total purity subject to coverage of at least \cvShFloor\%, over the
valid grid $\tabs\leq\tsel$. On the full benchmark this objective selects
$(\cvShFullOptPos,\cvShFullOptNeg)$ with \cvShFullOptPurity\% purity at
\cvShFullOptCoverage\% coverage, one abstention-grid step from the inherited
point and identical in purity (Table~\ref{tab:splithalf}). To quantify
threshold-selection optimism, we then split the \cvBenchN{} queries into two
halves at random, stratified by matchability, select the objective's optimum on
one half, and score it on the other; \cvShReps{} such splits scored in both
directions give \cvShEvals{} held-out evaluations (seed \cvShSeed). Selected
thresholds vary across \cvShDistinct{} grid points along the flat purity ridge
of Table~\ref{tab:grid}: $(\cvShModePos,\cvShModeNeg)$ is chosen most often
(\cvShModeShare\% of evaluations), and \cvShNearShare\% of selections fall
within $0.02$ (\tsel) and $0.05$ (\tabs) of the inherited point. Mean in-sample
purity of the selected point is \cvShInPurity\% against \cvShOutPurity\% held
out, an optimism of \cvShOptimism{} percentage points (2.5--97.5th percentile
$[\cvShOptimismLo,\cvShOptimismHi]$). On the same held-out halves the inherited
point scores \cvShInhOutPurity\% purity at \cvShInhOutCoverage\% coverage; the
objective-selected point differs from it by $\cvShDiffPurity$ points of purity
$[\cvShDiffPurityLo,\cvShDiffPurityHi]$ and $\cvShDiffCoverage$ points of
coverage $[\cvShDiffCoverageLo,\cvShDiffCoverageHi]$, buying volume with lower
positive purity (\cvShOutPosPurity\% versus \cvShInhOutPosPurity\%). Coverage
floors of 75\% and 85\% give the same picture (optimism \cvShAltSeventyFiveOptimism{}
and \cvShAltEightyFiveOptimism{} points). Two conclusions follow. Threshold-selection
optimism on this benchmark is small, and the inherited point is not an outlier
of the objective. The objective itself, however, under-determines the point:
total purity is flat across the ridge, so the choice among these points is a
coverage-versus-positive-purity decision that must be specified explicitly.
This check bounds only the threshold component of selection optimism; the
model and prompt configurations were also chosen on this benchmark and remain
in-sample, which is why fresh-data validation is planned
(Section~\ref{sec:discussion}).

\subsection{Residual Errors and Review}
\label{sec:residual}
\paragraph*{Shared false abstentions}
The \cvDualWrongNeg{} false no-match labels at the operating point remain
abstentions even at a zero confidence threshold. Of these,
\cvResidNativeNull{} already have null record IDs from both judges. Their
persistence is a limitation of thresholding these stored outputs, not proof of
irreducible task ambiguity. \cvResidGoldMulti{} have multi-record gold labels;
multiple acceptable records broaden the set of correct selections and do not
justify abstaining.

At the deployed gate, local GPT OSS 120B correctly selects on
\cvResidOssRecovers{} of these rows and Sonnet on \cvResidSonnetRecovers{}.
Their union covers \cvResidEitherRecovers{}, leaving \cvResidNoneRecovers{}
unresolved by all four stored configurations. This union is an oracle diagnostic,
not the performance of a deployable third-judge policy: no rule for deciding when
to trust the third judge has been evaluated. Shared errors may reflect reasoning,
prompt design, representation, or annotation; the counts do not identify a cause.

\paragraph*{Deferred rows and auditing}
The review set contains \cvDualReviewN{} rows, of which
\cvDualReviewMatchable{} are matchable. At the deployed gate, at least one
constituent is correct on \cvDualReviewEither{}; both are incorrect on
\cvDualReviewNeither{}. This measures available suggestions, not human review
accuracy or time saved. Crucially, the \cvDualWrongNeg{} shared false abstentions
are in the \emph{accepted negative set}, not in this review queue. Reviewing only
disagreements would miss them. A practical audit must sample accepted labels,
particularly negatives, as well as adjudicate deferred cases.

\section{Discussion and Limitations}
\label{sec:discussion}
\paragraph*{From retrospective curation to training}
% [R1-W1] Tied to the new check: says which part of the optimism is now bounded
% and which part is not.
Teacher and threshold selection reuse the evaluation benchmark, making these
retrospective estimates. The split-half check bounds the threshold-selection
component of that optimism at a fraction of a percentage point; it cannot bound
the optimism from choosing these two teachers and their prompts on the same
data, which only fresh queries can. Agreement favors easier examples and changes
class proportions. Students may inherit systematic false abstentions or perform
poorly on the deferred tail despite high label purity.

% [R1-W2] The reviewer notes that student utility is entirely future work. We
% do not claim it; the paragraph now states the planned protocol concretely
% (arms, controls, metrics, held-out data) so the scope of the missing evidence
% is explicit rather than implied.
In future work, we plan to curate fresh training queries with adjudicated
disagreements and audited negative labels, followed by LoRA fine-tuning of
GPT-OSS 20B and Gemma 4 12B. A controlled comparison should train students of
equal size on expert labels, single-teacher labels, dual-teacher labels, and a
positive-only or independently reviewed negative-label arm; control training
size; use several seeds; and evaluate \fhalf, \nfmr, latency, and cost on fresh
held-out queries with entity separation where feasible. The existing evaluation
benchmark must remain outside training. These are future experiments; the
present replay establishes neither student performance nor production cost
savings, and its purity figures should not be read as predictions of either.

\paragraph*{Scope and uncertainty}
% [R1-W3] Each of the reviewer's evidence limitations is now stated with its
% consequence, not only listed: seed repeats, prompt/demonstration confound,
% single task/provider/retriever, and the uncalibrated verbalized confidence.
This is one English-dominant task from one provider, conditional on a fixed
retriever, so the purity levels reported here should not be transferred to
other catalogs or retrievers without re-measurement. There is one stored output
per configuration--query, without independent seed repeats; the bootstrap
intervals therefore capture query sampling, not decoding variability, and a
rerun of either judge could move individual retention decisions. Differing
prompts and demonstrations prevent attribution of the observed differences to
model family alone. Agreement does not imply independence: both judges see the
same candidate pool, and the shared false abstentions in
Section~\ref{sec:residual} show that their errors co-occur. Verbalized name
confidence is not a calibrated probability; the gates act on it only as an
ordinal score, which is why Fig.~\ref{fig:sweep} reports purity as a function
of the threshold rather than assuming purity equals confidence. The intervals
condition on the observed configurations and rows; they omit configuration
selection (Section~\ref{sec:splithalf} bounds the threshold part only),
possible entity clustering, temporal drift, and uncertainty in expert labels.

\paragraph*{Data governance and reproducibility}
The benchmark schema has no cardholder identity, account number, transaction
amount, or timestamp. Merchant information can nonetheless be sensitive.
The proposed curation inference uses local endpoints, while the historical
reference and demonstration-construction workflows included hosted models;
local curation does not imply an entirely local provenance chain.
% [R1-W5] The reviewer flags non-release of data, code and prompts. The
% restriction cannot be lifted here, so the paragraph now says exactly what
% IS reproducible: the replay is deterministic from stored outputs, every
% number comes from one script, and the protocols (gates, sweep, split-half)
% are fully specified so they can be re-run on other benchmarks.
The analysis is a deterministic replay of the stored judge outputs: one script
produces every table, figure, macro, and input hash in this paper, and the
split-half protocol is specified completely in Section~\ref{sec:splithalf} so
that it can be applied to other benchmarks. Due to corporate confidentiality
restrictions, we cannot publicly release the dataset, implementation code, or
prompt templates. We describe the decision rules, evaluation protocol, and
aggregate results to support methodological transparency.

\section{Conclusion}
Separately gated agreement retains \cvDualTotalN{} of \cvBenchN{} queries at
\cvDualTotalPurity\% purity in this retrospective study. Ordered thresholds
prevent contradictory labels, but the ablation and \cvDualWrongNeg{} persistent
false abstentions show why accepted negatives need independent auditing.
% [R1-W1][R1-W4] One sentence on the held-out check.
Under a stated objective, the inherited operating point coincides with the
full-benchmark optimum within one grid step, and threshold-selection optimism
on held-out halves is \cvShOptimism{} percentage points; configuration
selection remains in-sample.
Higher reasoning effort yields no clear \fhalf{} gain at
$\cvEffLatencyMinX$--$\cvEffLatencyMaxX\times$ median latency in the matched
comparisons, supporting low effort here without a general claim about disabling
reasoning. Fresh-data curation, LoRA fine-tuning, and independent student
evaluation remain necessary to establish downstream value.

\section*{Acknowledgment}
We thank Tom\'a\v{s} Drietomsk\'y and Vojtech Pacak of Economic Intelligence,
Mastercard, Czech Republic, for their valuable collaboration and for sharing
the data, prompts, and original evaluation code.
This research was supported by the Singapore Ministry of Education (MOE) Academic Research Fund (AcRF) Tier~1 Grant
(Grant No.~24-SIS-SMU-131).

\bibliographystyle{IEEEtran}
\bibliography{references}
\end{document}

%% file: data/curation_values.tex
\newcommand{\cvOpPosTau}{0.86}
\newcommand{\cvOpNegTau}{0.80}
\newcommand{\cvDeployedTau}{0.70}

\newcommand{\cvDualPosN}{938}
\newcommand{\cvDualPosPurity}{99.47}
\newcommand{\cvDualPosCi}{[98.76,99.77]}
\newcommand{\cvDualNegN}{695}
\newcommand{\cvDualNegPurity}{93.38}
\newcommand{\cvDualNegCi}{[91.28,95.00]}
\newcommand{\cvDualWrongNeg}{46}
\newcommand{\cvDualTotalN}{1{,}633}
\newcommand{\cvDualTotalPurity}{96.88}
\newcommand{\cvDualTotalCi}{[95.92,97.62]}
\newcommand{\cvDualCoverage}{81.7}
\newcommand{\cvDualReviewN}{367}

\newcommand{\cvDualReviewMatchable}{322}
\newcommand{\cvDualReviewEither}{358}
\newcommand{\cvDualReviewNeither}{9}
\newcommand{\cvDualPosErrors}{5}

\newcommand{\cvMuseWrongNegAtOp}{88}
\newcommand{\cvGemmaWrongNegAtOp}{69}

\newcommand{\cvMuseTotalPurityAtOp}{94.81}
\newcommand{\cvGemmaTotalPurityAtOp}{94.63}

\newcommand{\cvSonnetTotalPurityAtOp}{96.57}
\newcommand{\cvMuseTotalNAtOp}{1{,}850}
\newcommand{\cvGemmaTotalNAtOp}{1{,}900}

\newcommand{\cvSonnetTotalNAtOp}{1{,}983}

\newcommand{\cvDualPosPurityAtSeventy}{98.91}
\newcommand{\cvDualPosNAtSeventy}{1{,}100}

\newcommand{\cvDualPosPurityAtNinetyFive}{99.79}
\newcommand{\cvDualPosNAtNinetyFive}{467}

\newcommand{\cvDualNegPurityAtEighty}{93.38}

\newcommand{\cvDualNegPurityAtNinety}{88.55}

\newcommand{\cvDualWrongNegAtNinety}{87}

\newcommand{\cvDualWrongNegMax}{1{,}265}

\newcommand{\cvResidOssRecovers}{14}
\newcommand{\cvResidSonnetRecovers}{16}
\newcommand{\cvResidEitherRecovers}{22}
\newcommand{\cvResidNoneRecovers}{24}
\newcommand{\cvResidGoldMulti}{31}
\newcommand{\cvResidNativeNull}{43}

\newcommand{\cvSymHighExtraNeg}{64}
\newcommand{\cvSymHighExtraWrongNeg}{40}
\newcommand{\cvSymHighPurity}{94.64}
\newcommand{\cvSymLowPosLoss}{162}
\newcommand{\cvSymLowPosErrorsRemoved}{7}
\newcommand{\cvBootstrapN}{4{,}000}
\newcommand{\cvMusePurityDelta}{2.07}
\newcommand{\cvGemmaPurityDelta}{2.25}

\newcommand{\cvSonnetPurityDelta}{0.31}
\newcommand{\cvMusePurityDeltaCi}{[1.40,2.78]}
\newcommand{\cvGemmaPurityDeltaCi}{[1.54,3.01]}

\newcommand{\cvSonnetPurityDeltaCi}{[-0.43,1.06]}
\newcommand{\cvMuseBestPurity}{94.86}
\newcommand{\cvGemmaBestPurity}{94.93}

\newcommand{\cvMuseBestN}{1{,}847}
\newcommand{\cvGemmaBestN}{1{,}715}

\newcommand{\cvInvertedOverlap}{19}

\newcommand{\cvBenchN}{2{,}000}
\newcommand{\cvBenchNeg}{696}
\newcommand{\cvBenchPos}{1{,}304}
\newcommand{\cvTauGridLo}{0.60}
\newcommand{\cvTauGridHi}{0.98}

\newcommand{\cvEffMuseMediumDelta}{-0.42}

\newcommand{\cvEffOssMediumDelta}{-0.15}

\newcommand{\cvEffOssHighDelta}{-0.50}

\newcommand{\cvEffSonnetMediumDelta}{-0.31}

\newcommand{\cvEffSonnetHighDelta}{+0.04}
\newcommand{\cvEffSonnetHighCi}{[-0.52,0.63]}
\newcommand{\cvEffSonnetHighLatencyX}{2.7}
\newcommand{\cvEffContrastN}{5}

\newcommand{\cvEffBootstrapN}{4{,}000}
\newcommand{\cvEffLatencyMinX}{1.8}
\newcommand{\cvEffLatencyMaxX}{5.0}

%% file: data/split_half_values.tex
\newcommand{\cvShReps}{200}
\newcommand{\cvShEvals}{400}
\newcommand{\cvShSeed}{42}
\newcommand{\cvShFloor}{80}
\newcommand{\cvShModePos}{0.86}
\newcommand{\cvShModeNeg}{0.81}
\newcommand{\cvShModeShare}{33}
\newcommand{\cvShDistinct}{16}

\newcommand{\cvShNearShare}{39}

\newcommand{\cvShInPurity}{96.93}
\newcommand{\cvShOutPurity}{96.79}

\newcommand{\cvShOptimism}{0.14}
\newcommand{\cvShOptimismLo}{-1.61}
\newcommand{\cvShOptimismHi}{1.80}

\newcommand{\cvShOutPosPurity}{99.14}

\newcommand{\cvShInhOutPurity}{96.88}

\newcommand{\cvShInhOutCoverage}{81.7}
\newcommand{\cvShInhOutPosPurity}{99.47}

\newcommand{\cvShDiffPurity}{-0.09}
\newcommand{\cvShDiffPurityLo}{-0.36}
\newcommand{\cvShDiffPurityHi}{0.00}
\newcommand{\cvShDiffCoverage}{+3.9}
\newcommand{\cvShDiffCoverageLo}{-0.2}
\newcommand{\cvShDiffCoverageHi}{8.6}
\newcommand{\cvShFullOptPos}{0.86}
\newcommand{\cvShFullOptNeg}{0.81}
\newcommand{\cvShFullOptPurity}{96.88}
\newcommand{\cvShFullOptCoverage}{81.8}

\newcommand{\cvShAltSeventyFiveOptimism}{0.14}

\newcommand{\cvShAltEightyFiveOptimism}{0.09}

%% file: data/table_judges.tex
\begin{table*}[t]
\centering
\caption{Judges on the frozen 2{,}000-row benchmark ($\tau{=}.70$). Quality metrics are percentages. Low/medium are provider-specific effort settings; off denotes disabled thinking. Time is median serial model-call latency (local workstation or Sonnet API), not production throughput or cost.}
\label{tab:judges}
\small
\setlength{\tabcolsep}{4pt}
\begin{tabular}{@{}llrrrrr@{}}
\toprule
Judge & Configuration & $F_{0.5}$ & $P$ & $R_m$ & N-FMR & s/row \\
\midrule
GPT OSS 120B & deployed, medium & 94.83 & 95.94 & 90.64 & 6.03 & -- \\
\textbf{Muse Glimmer 30B} & zero-shot, low, v1 & 96.76 & 97.89 & 92.48 & 2.16 & 35.2 \\
\textbf{Gemma 4 31B} & $k{=}20$, off, v2 & 95.27 & 95.69 & 93.63 & 5.75 & 22.2 \\
GPT OSS 120B & low, v1 (local) & 94.95 & 95.89 & 91.33 & 5.60 & 16.9 \\
Claude Sonnet 4.5 & low, v1 (API) & 97.41 & 97.81 & 95.86 & 2.59 & 14.2 \\
\bottomrule
\end{tabular}
\end{table*}

%% file: data/table_operating.tex
\begin{table*}[t]
\centering
\caption{Curated labels at $\tau_{\mathrm{sel}}{=}0.86$, $\tau_{\mathrm{abs}}{=}0.80$. Purity and coverage are percentages; review is the remainder of the 2{,}000 rows. The policies select different subsets, so this comparison does not hold coverage fixed.}
\label{tab:operating}
\small
\setlength{\tabcolsep}{3.5pt}
\begin{tabular}{@{}lrrrrrrrr@{}}
\toprule
Curator & Positives & Purity & Negatives & Purity & Total & Purity & Coverage & Review \\
\midrule
Muse Glimmer 30B & 1{,}077 & 99.26 & 773 & 88.62 & 1{,}850 & 94.81 & 92.5 & 150 \\
Gemma 4 31B & 1{,}175 & 97.19 & 725 & 90.48 & 1{,}900 & 94.63 & 95.0 & 100 \\
GPT OSS 120B & 1{,}212 & 96.78 & 765 & 86.41 & 1{,}977 & 92.77 & 98.8 & 23 \\
Claude Sonnet 4.5 & 1{,}257 & 98.25 & 726 & 93.66 & 1{,}983 & 96.57 & 99.2 & 17 \\
\midrule
\textbf{Muse $\times$ Gemma} & 938 & \textbf{99.47} & 695 & 93.38 & 1{,}633 & \textbf{96.88} & 81.7 & 367 \\
\bottomrule
\end{tabular}
\end{table*}

%% file: data/table_ablation.tex
\begin{table}[t]
\centering
\caption{Gate ablation with fixed outputs. Asymmetric uses $(.86,.80)$; positive-only uses $\tau_{\mathrm{sel}}=.86$. Err.$+$ and Err.$-$ count incorrect labels; purity and coverage are percentages.}
\label{tab:ablation}
\small
\setlength{\tabcolsep}{3pt}
\begin{tabular}{@{}lrrrr@{}}
\toprule
Policy & Err.$+$ & Err.$-$ & Purity & Coverage \\
\midrule
Symmetric .70 & 12 & 46 & 96.77 & 89.7 \\
Symmetric .80 & 11 & 46 & 96.82 & 89.7 \\
Symmetric .86 & 5 & 86 & 94.64 & 84.8 \\
Asymmetric & 5 & 46 & 96.88 & 81.7 \\
Positive only & 5 & 0 & 99.47 & 46.9 \\
\bottomrule
\end{tabular}
\end{table}

%% file: data/table_grid.tex
\begin{table}[t]
\centering
\caption{Total label purity (\%) over the gate grid. The excluded cell violates the protocol's disjointness guarantee and overlaps on \cvInvertedOverlap{} rows. Bold marks the illustrative operating point; polarity-specific quantities are separable, but total purity also depends on label counts.}
\label{tab:grid}
\footnotesize
\setlength{\tabcolsep}{4pt}
\begin{tabular}{@{}lrrrr@{}}
\toprule
$\tau_{\mathrm{sel}}$ $\backslash$ $\tau_{\mathrm{abs}}$ & 0.70 & 0.75 & 0.80 & 0.86 \\
\midrule
0.80 & 96.82 & 96.82 & 96.82 & --- \\
0.86 & 96.87 & 96.88 & \textbf{96.88} & 94.64 \\
0.90 & 96.83 & 96.84 & 96.84 & 94.57 \\
0.95 & 95.95 & 95.95 & 95.96 & 92.90 \\
\bottomrule
\end{tabular}
\end{table}

%% file: data/table_split_half.tex
\begin{table}[t]
\centering\footnotesize\setlength{\tabcolsep}{2pt}
\caption{Held-out check of the operating point. The stated objective maximizes total purity subject to coverage $\geq$ 80\% over the valid grid. Held-out rows average 400 evaluations from 200 stratified random half-splits (seed 42): thresholds are selected on one half and scored on the other. This quantifies threshold-selection optimism only; model and prompt selection also reused this benchmark. Purity and coverage are percentages.}
\label{tab:splithalf}
\begin{tabular}{@{}lccrrrr@{}}
\toprule
Policy & \tsel & \tabs & Total & Pos. & Neg. & Cov. \\
\midrule
Inherited (full) & 0.86 & 0.80 & 96.88 & 99.47 & 93.38 & 81.7 \\
Objective optimum (full) & 0.86 & 0.81 & 96.88 & 99.47 & 93.40 & 81.8 \\
Objective-selected (held-out) & 0.60--0.86 & 0.60--0.81 & 96.79 & 99.14 & 93.39 & 85.5 \\
Inherited (held-out) & 0.86 & 0.80 & 96.88 & 99.47 & 93.39 & 81.7 \\
\bottomrule
\end{tabular}
\end{table}